\documentclass[12pt, letterpaper]{article}

\usepackage{framed,multirow}

\usepackage{amsmath,amsfonts,amssymb}
\usepackage{latexsym}
\usepackage{amssymb}
\usepackage{booktabs}
\usepackage[export]{adjustbox}
\usepackage{url}
\usepackage{xcolor}
\usepackage{todonotes}
\definecolor{newcolor}{rgb}{.8,.349,.1}

\usepackage{hyperref}
\usepackage[capitalize]{cleveref}
\crefname{section}{Sec.}{Secs.}
\Crefname{section}{Section}{Sections}
\Crefname{table}{Table}{Tables}
\crefname{table}{Tab.}{Tabs.}
\def\etal{\textit{et al.}}

\usepackage{pifont}

\usepackage[switch,pagewise]{lineno} 

\newtheorem{theorem}{Theorem}[section]
\newtheorem{definition}{Definition}[section]

\title{Toward Mesh-Invariant 3D Generative Deep Learning with Geometric Measures}

\author{Thomas Besnier$^1$, Sylvain Arguill\`ere$^2$ \\ Emery Pierson $^1$ $^, $ $^5$, Mohamed Daoudi$^3$ $^,$ $^4$}
\date{%
    $^1$Univ. Lille, CNRS, Centrale Lille, UMR 9189 CRIStAL, Lille, F-59000, France\\%
    $^2$Univ. Lille, CNRS, UMR 8524 - Laboratoire Paul Painlev\'e, F-59000 Lille, France \\
     $^3$Univ. Lille, CNRS, Centrale Lille, Institut Mines-Télécom, UMR 9189 CRIStAL, Lille, F-59000, France\\
     $^4$IMT Nord Europe, Institut Mines-Télécom, Univ. Lille, Centre for Digital Systems, Lille, F-59000, France\\
     $^5$Universität Wien, Austria\\[2ex]
   \today
}

\begin{document}

\maketitle

\begin{abstract}
3D generative modeling is accelerating as the technology allowing the capture of geometric data is developing. However, the acquired data is often inconsistent, resulting in unregistered meshes or point clouds. Many generative learning algorithms require correspondence between each point when comparing the predicted shape and the target shape. We propose an architecture able to cope with different parameterizations, even during the training phase. In particular, our loss function is built upon a kernel-based metric over a representation of meshes using geometric measures such as currents and varifolds. The latter allows to implement an efficient dissimilarity measure with many desirable properties such as robustness to resampling of the mesh or point cloud. We demonstrate the efficiency and resilience of our model with a generative learning task of human faces.
\end{abstract}

\section{Introduction}

\begin{figure}
    \centering
    \includegraphics[width=0.8\linewidth]{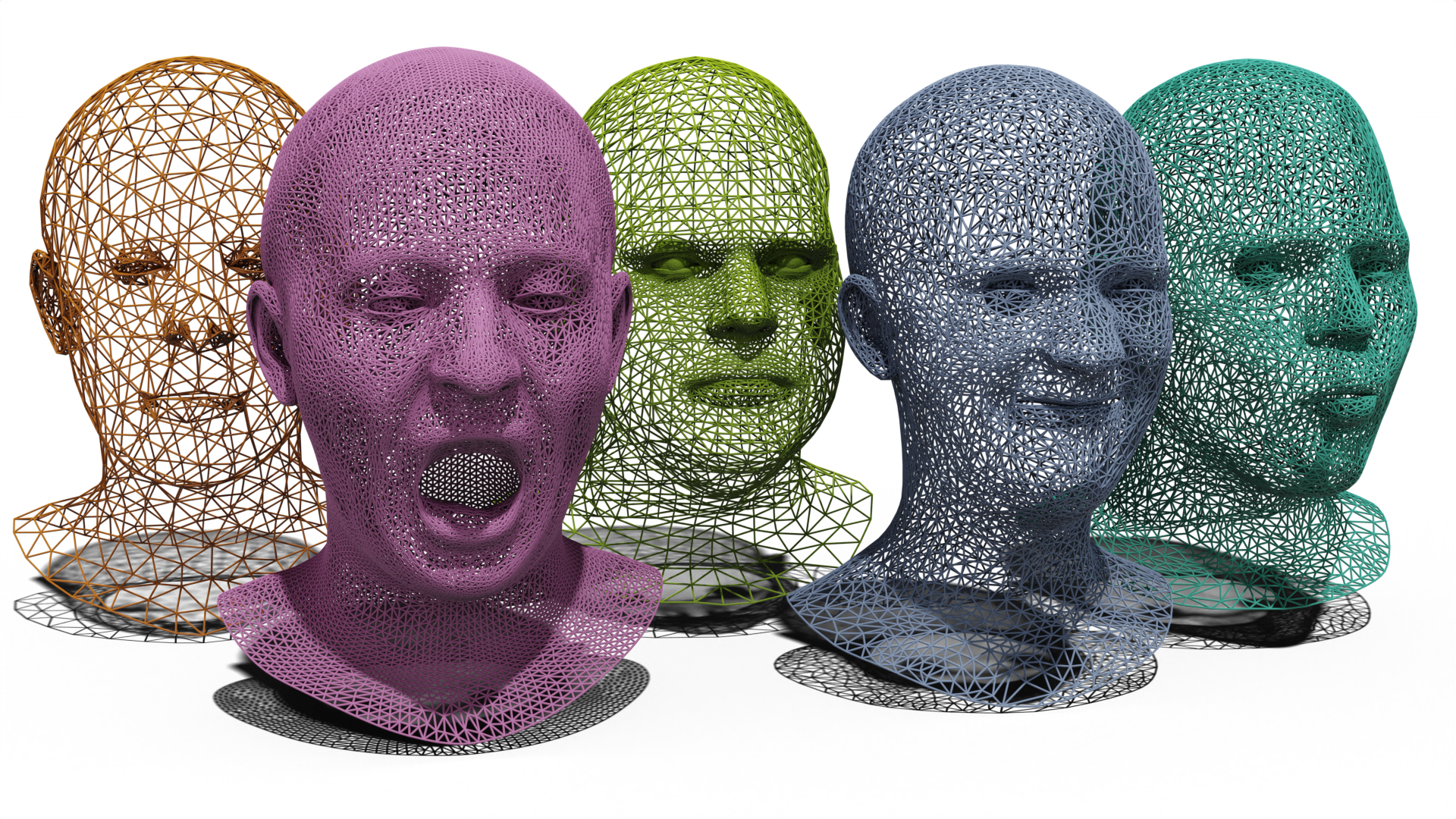}
    \caption{Several reparameterized meshes from the COMA dataset \cite{COMA}. Captured geometric data does not have identical graph structure and correspondence between points. Even the resolution driven by the number of points/vertices is subject to variations along a dataset of scans.}
    \label{fig:intro_figure}
\end{figure}

In this paper, we focus on the generation of believable deformations of 3D faces, which has practical applications in various graphics fields, including 3D face design, augmented and virtual reality, as well as computer games and animated films. Despite the rapid progress in 3D face generation thanks to deep learning, existing methods have not yet been able to learn from unregistered scans with varying parameterizations (see~\Cref{fig:intro_figure}). 
Indeed, one major restriction of the current methods such as graph convolutional networks \cite{bouritsas_neural3DMM_2019, COMA} for generating these facial deformations is their reliance on a unified graph structure (required by the network architecture), along with point correspondence (for the loss function) between the target and the predicted mesh. This is especially problematic when handling real-world data: surfaces can be acquired with different technologies (LIDAR, 3D scans, Neural radiance fields \cite{mildenhall2020nerf}, ...) which results in inconsistent and heterogeneous databases \cite{BU-3DFE, Bosphorus, Texas3D}. 

Therefore, costly registration algorithms are often needed to create consistent datasets, and can even require manual intervention. 
As the registration of large point clouds or meshes can require hours or even days of processing time, especially with the growing number of available databases, this has motivated ongoing research on efficient methods and hardware acceleration techniques to make them more practical for real-world applications. 

To address these issues, we propose an auto-encoder architecture that can, by design, be trained on inconsistent datasets with no unifying graph structure, resolution or point correspondence. Moreover, our approach works directly with explicit surface meshes, avoiding complex computation with volumetric data or functional description of the shapes.
The main tools employed in our approach consist of a PointNet \cite{Charles_PointNet_2017} encoder with the ability to process point clouds with variable number of points and map them to a low dimensional latent space, alongside a novel loss function that can withstand variations in parameterization. In particular, the proposed loss is robust to resampling of the mesh thanks to a representation of shapes in terms of geometric measures such as varifolds.
Recently, by using a kernel metric on the space of these varifolds, Kaltenmark~\etal~\cite{kaltenmark2017general} have proposed a general framework for 2D and 3D shape similarity measures, invariant to parameterization and equivariant to rigid transformations. This framework showed state-of-the-art performances in different shape matching frameworks, which motivates our approach. To the best our knowledge, this is the first use of kernel metric in the space of geometric measures as cost function in deep learning.

Our results show that the presented model is indeed able to learn on meshes with different parameterization. Moreover, our learned auto-encoder demonstrates expressive capabilities to rapidly perform interpolations, extrapolations and expression transfer through the latent space. Our main contributions are as follow:

\begin{itemize}
    \item  We propose a generative learning method using a parameterization invariant metric based on geometric measure theory. More precisely, we use kernel metrics on varifolds, with a novel multi-resolution kernel. We use it as a dissimilarity measure between the generated mesh and the target mesh during training. We highlight in particular many desirable properties of this metric compared to other metrics used in unsupervised geometric deep learning.
    \item We propose a robust training method for face registration. It is composed of an asymmetric auto-encoding architecture, that allows to learn efficiently on human faces, with a loss function based on a varifold multi-resolution metric. This approach allows us to learn on inconsistent databases with no correspondence between vertices. Moreover, we are able to learn efficiently an expressive latent space.
    \item To validate the robustness of our approach, we conduct several experiments, using the trained model, including face generation, interpolation, extrapolation and expression transfer.
\end{itemize}
\section{Related work}
\subsection{Geometric deep learning on meshes}
Within the field of geometric deep learning, a critical challenge is generalizing operators such as convolution and pooling to meshes.
The first idea was a direct application of convolutions on the well-suited transformation of 3D meshes, such as multi-view images~\cite{qi2016volumetric}, or 3D (Convolutional neural networks) CNNs on volumetric data~\cite{gadelha20173d}. 
Recent advances have shown that the latter, with other representations such as signed distance or radiance fields can be used in 3D CNNs and allow the reconstruction of 3D shapes. It was successfully used in applications such as geometric aware image generation~\cite{StyleGAN3D}, shape reconstruction from images~\cite{Chancvpr2022}, or partial shape completion~\cite{chibane2020implicit}. However, these approaches remain expensive and time-consuming to obtain detailed shapes, making them unpractical for generative tasks compared to mesh-based approaches~\cite{Dundar2022}.

By proposing a permutation invariant approach to learning on point clouds, PointNet~\cite{Charles_PointNet_2017} has opened a new and practical way to easily encode information of 3D points. Several improvements have been applied over the years with multi-scale aggregation techniques, such as PointNet++~\cite{qi2017pointnet++}, KPConv~\cite{thomas2019kpconv}, or the PointTransformer~\cite{zhao2021point}, and recently, PointNext~\cite{qian2022pointnext} has shown that such an approach can scale on large databases. However, while PointNet showed provable robustness to different discretization, PointNet++, and its derivatives are based on fixed-size neighborhood aggregation, thus being sensitive to the mesh discretization. 

In the meantime, surface-based filters have been proposed to improve results and take in account the geometry of the surface. The first filters propose to exploit the graph structure of surface meshes~\cite{masci2015geodesic}, and apply Graph Neural Networks (GNN) on the underlying graph. The advantage is that it becomes easy to generalize popular CNNs architecture like autoencoders or U-Net. However, original GNN lacks expressivity because of their isotropy (they gather information in all directions equally), and several anisotropic filters have been proposed~\cite{de2020gauge, mitchel2021field, sharp2022diffusionnet} and shown state-of-the-art results on shape classification and segmentation. Whether or not they can be applied to generative models remains however an open question.
Recently, Lemeunier et al.~\cite{lemeunier2022representation} proposed to work in the spectral domain to learn on human bodies, but this approach needs the meshes to be converted to the spectral domain during learning and it is unclear how to adapt it to unparameterized data. Neural3DMM~\cite{bouritsas_neural3DMM_2019}, which is based on the spiral ordering of mesh neighborhoods, has shown its efficiency on 3D faces~\cite{Bahri_SMF_2021, otberdout2022sparse}.

\begin{figure}[h!]
    \centering
    \includegraphics[width=0.95\linewidth]{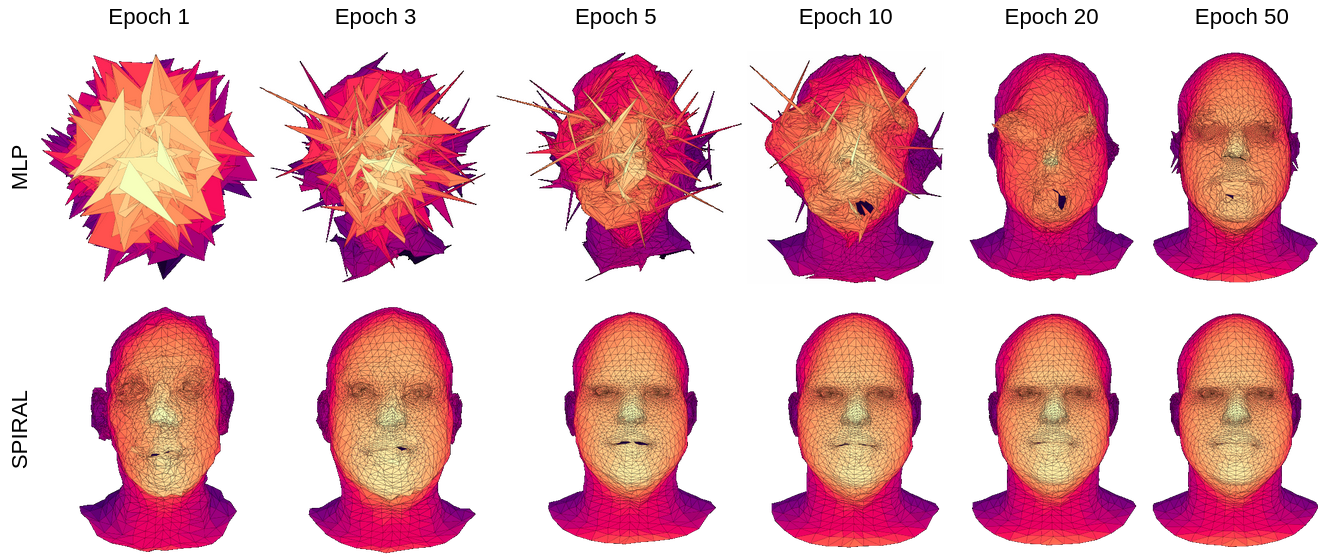}
    \caption{Visualization of the learning process of a single mesh using our model with different decoders. The first row shows the results using a MLP as the decoder and the for the second row, we used a spiral convolution. In addition to a slower learning process, the MLP struggles to map a linear space (the latent space) to a high-dimensional and non-linear one.}
    \label{fig:compare_decoder}
\end{figure}

In this work, we follow recent advances~\cite{Bahri_SMF_2021} and we propose an asymmetric autoencoder, with a PointNet architecture for computing the latent vector from an unregistered mesh: the simplicity of PointNet combined to its proven robustness makes it the ideal candidate, as opposed to more expressive, but less parameterization robust approaches. In the contrary, the decoder is made of a template-dependent architecture, namely SpiralNet, for two reasons: the proven results on human faces, and the fact that spiral convolutions incorporate a better prior on the deformation model. We illustrate this with a very simple experiment in which a model learn a Multi-Layer Perceptron (MLP) and an SpiralNet decoder on the task of mapping a single vector to a target face. We observe in~\Cref{fig:compare_decoder} that the target shape is reached faster, and the intermediate shape are representing human faces, as opposed to the MLP decoder. This property will allow our model to reach more easily a suitable registration of shapes.

\subsection{Robust generative learning in 3D}

In the literature, "robustness" is often a welcomed side property of the model \cite{Hanocka_MeshCNN_2019, sharp2022diffusionnet, Charles_PointNet_2017}, but these models still rely on a consistent database for the training phase.\\
In 3D machine learning, while supervised learning relies on training data that has a precise and consistent order, unsupervised learning operates without any prior correspondence and instead utilizes self-organization to model the inherent geometry of the data.
Here, we explore tools to allow \textbf{complete unsupervised} 3D generative deep learning. Little work has been done on this subject, especially for generative tasks, as in \cite{groueix_3DCODED_2018} or \cite{Croquet_Diff_Reg_OT_2021}. Most of the time, unsupervised learning tasks are performed by considering a functional representation of shapes \cite{Eisenberger_DeepShells_2020, Roufosse_SURFMNet_2019, Cao_Bernard_2022} but these methods have not been applied for face generation yet. We also mention \cite{Liu_2019,Bahri_SMF_2021} who proposed an hybrid supervised/unsupervised learning protocols. Most of the other unsupervised registration methods are computed with non learning method such as LDDMM (Large deformation diffeomorphic metric mapping) \cite{Beg_Miller_Trouve_Younes_LDDMM_2005} or elastic shape matching \cite{Hartman_H2Match_2023} with a recent exception in \cite{Croquet_Diff_Reg_OT_2021} where the authors learn an unsupervised diffeomorphic registration with an auto-encoder architecture that uses optimal transport for the loss function.

\subsection{Geometric loss functions}
In the following, we consider 3D datasets with elements being meshes. We describe such a mesh $X = \{V(X), E(X), F(X)\}$ with $V(X) = (x_i)_i \in \mathbb{R}^{3\times n_X}$ the set of vertices of $X$, $E(X)$ its edges and $F(X)$ its faces with $n_X = |V(X)|$ being the resolution of the mesh. Following this, $\hat{X}$ will refer to a reconstruction of $X$.

As mentioned in the introduction, our goal is to learn explicit 3D data on inconsistent databases. In real-world situations, without prior registration, we usually lack the correspondence between points, making it difficult to compute the mean squared error (MSE), or any similar euclidean distance. As a result, we investigate dissimilarity metrics that are robust or invariant to the parameterisation with a relevant geometric meaning. We review some conventional distances, some of them will be used to evaluate our method:

\begin{itemize}
    \item The Mean-squared error metric requires correspondence between each points
    \begin{equation}
        \mathcal{L}^{MSE}(X, \hat{X}) = \frac{1}{n_X}\sum_{x \in X} \| x - \hat{x}\|_2^2
    \end{equation}
    It is the most commonly used dissimilarity measure for supervised learning tasks.
    \item The Hausdorff distance (H) is a strong metric with a powerful ability to generalize to heterogeneous spaces \cite{ROTE1991123}.
    \begin{equation}
        \mathcal{L}^H(X,\hat{X}) = \max \left\{ \sup_{x \in X}d(x,\hat{X}), \sup_{\hat{x} \in \hat{X}}d(X,\hat{x}) \right\}
    \end{equation}
    where $d$ is a pre-defined distance, usually euclidean.\\
    Unfortunately, optimizing with respect to this metric will imply to correct only one point at each step of the gradient which make it an inefficient loss function.
    \item The Chamfer distance (CD) \cite{wu2021densityaware} is strongly linked with the iterative closest point algorithm (ICP) as it is basically the objective function to minimize for this algorithm:
    \begin{multline}
        \mathcal{L}^{CD}(X, \hat{X}) = \frac{1}{n_X}\sum_{\hat{x} \in             V(\hat{X})}\min_{x \in V(X)}\|\hat{x} - x \|_2^2 \\ + \frac{1}{n_{\hat{X}}}\sum_{x \in V(X)}\min_{\hat{x} \in V(\hat{X})}\|\hat{x} - x \|_2^2
    \end{multline}
In fact, as a loss function we only require the directed Chamfer distance (DCD) given by the first term in the previous expression
 \begin{equation}
        \mathcal{L}^{DCD}(X, \hat{X}) = \frac{1}{n_X}\sum_{\hat{x} \in             V(\hat{X})}\min_{x \in V(X)}\|\hat{x} - x \|_2^2
    \end{equation}
This loss is notably used for existing unsupervised learning tasks in \cite{groueix_3DCODED_2018, Liu_2019,Bahri_SMF_2021, Chen_DeepUL_2019}. However, the Chamfer distance can suffer from poor performances for at least two reasons: the first one is that the use of $\min$ operator makes the loss unstable, because it is not fully differentiable with respect to a mesh positions. Second, it can be sensitive to outliers or collapse on points of a mesh. Generally, this loss is regularized using an additional term constraining the mesh deformation during the training phase. Various techniques exists, such as adding an edge loss $\mathcal{L}^{edges}$ with respect to a template mesh $X^t$. It can also be combined with a Laplacian loss to ensure the smoothness of the reconstructed mesh. 
    
\item The Wasserstein distance (also called the \textit{earth-mover distance}) from optimal transport theory is very popular to compute a distance between shapes. However, solving the optimal transport problem is NP-hard~\cite{peyreOT}, and the distance can be approximated with the debiased Sinkhorn divergence (SD) developed in \cite{Cuturi_sinkhorn_2013} and recently used in \cite{Feydy_SD_2019}. \\
To compute this distance, we represent a mesh $X$ as an aggregation of Dirac's measure $A(X) = \sum_i a^X_i \delta_{c^X_i}$ which gives a sum representing the center of faces $(c^X_i)_i$, weighted by their corresponding area $(a^X_i)_i$.
    \begin{multline}
        \mathcal{L}^{SD}(X, \hat{X}) = OT_{\epsilon}(A(X),A(\hat{X})) - \frac{1}{2}OT_{\epsilon}(A(X),A(X)) \\ - \frac{1}{2}OT_{\epsilon}(A(\hat{X}),A(\hat{X}))
    \end{multline}
with 
    \begin{multline*}
        OT_{\epsilon}(\alpha,\beta) = \min_{\pi \in \Pi}\sum_{i=1}^N\sum_{j=1}^M\pi_{i,j}\frac{1}{p}\|c^X_i - c^{\hat{X}}_j\|_p + \epsilon KL(\pi \| \alpha \otimes \beta)
    \end{multline*}
where $\pi$ is the regularized transport plan. The Kullback-Leibler divergence (KL) is a regularization term, called entropic penalty, and the blur $\epsilon$ is a hyperparameter that indicates how strong is the approximation of the Wasserstein distance with $OT_{\epsilon} \xrightarrow[\epsilon \rightarrow 0]{} OT$ the true Wasserstein distance. This hyperparameter needs to be adapted to each different task. Moreover, we observed in our experiments that the computation and differentiation of the distance can still be expensive, and makes it unpractical for learning on large datasets.
\end{itemize}

In contrast to all methods, the varifold approach is built using reproducing kernels, and thus have a spatial support: the outliers are not seen by the loss. Moreover, by taking into account every relationship in each pair of points, it is fully differentiable with respect to mesh position. Finally, the use of normals in the loss, allows to account for the shape of the surface, instead of seeing a mesh as just a cloud of points of $\mathbb{R}^3$. We denote our proposed loss $\mathcal{L}^{GM}$ and summarize its advantages in~\Cref{tab:comparison_metrics}.

\begin{table}[ht!]
\centering
\begin{tabular}{l|lllll}
Property / Loss         & \multicolumn{1}{c}{$\mathcal{L}^{MSE}$} & \multicolumn{1}{c}{$\mathcal{L}^H$} & \multicolumn{1}{c}{$\mathcal{L}^{DCD}$} & $\mathcal{L}^{SD}$   & \multicolumn{1}{c}{$\mathcal{L}^{GM}$} \\ 
\hline
Unsupervised        & \multicolumn{1}{c|}{\ding{55}}   & \multicolumn{1}{c|}{\ding{51}}  & \multicolumn{1}{c|}{\ding{51}}       & \multicolumn{1}{c|}{\ding{51}} & \multicolumn{1}{c|}{\ding{51}}        \\
Smooth gradient  & \multicolumn{1}{c|}{\ding{51}}     & \multicolumn{1}{c|}{\ding{55}}  & \multicolumn{1}{c|}{\ding{55}}       & \multicolumn{1}{c|}{\ding{51}} & \multicolumn{1}{c|}{\ding{51}}        \\
Position   & \multicolumn{1}{c|}{\ding{51}}   & \multicolumn{1}{c|}{\ding{51}}  & \multicolumn{1}{c|}{\ding{51}}       & \multicolumn{1}{c|}{\ding{51}} & \multicolumn{1}{c|}{\ding{51}}        \\
Orientation & \multicolumn{1}{c|}{\ding{55}}  & \multicolumn{1}{c|}{\ding{55}}  & \multicolumn{1}{c|}{\ding{55}}       & \multicolumn{1}{c|}{\ding{55}} & \multicolumn{1}{c|}{\ding{51}}        \\
Tunable      & \multicolumn{1}{c|}{\ding{55}} & \multicolumn{1}{c|}{\ding{55}}  & \multicolumn{1}{c|}{\ding{55}}       & \multicolumn{1}{c|}{\ding{51}} & \multicolumn{1}{c|}{\ding{51}}       
\end{tabular}
\caption{Summary of the properties of the aforementioned dissimilarity metrics}
\label{tab:comparison_metrics}
\end{table}

As we will see of the experiments, we do not need any regularization to obtain plausible mesh as outputs of our method.
\section{Our approach }
\begin{figure*}[t!]
    \centering
    \includegraphics[width=0.95\linewidth]{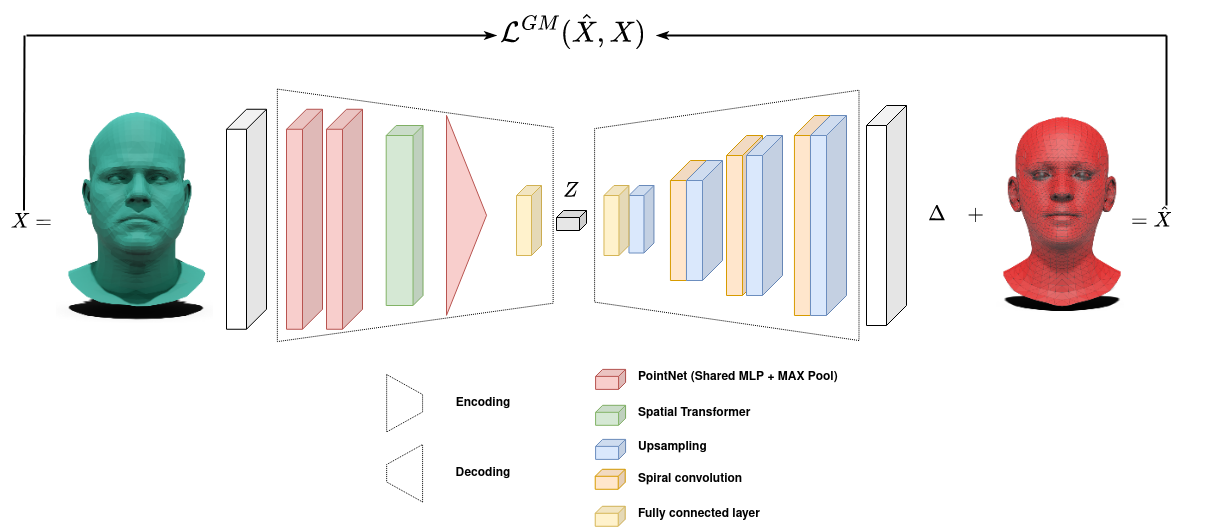}
    \caption{Architecture overview. The encoder takes a mesh $X$ of any parameterization as input and outputs a deformation $\Delta$ added to a chosen template to obtain a registered mesh $\hat{X}$ with a similar topology as the template.}
    \label{fig:autoencoder}
\end{figure*}

To complete the face generation task, we propose a simple auto-encoder detailed in~\Cref{fig:autoencoder}.
The main originality of our model comes from the loss function based on a representation of meshes with discrete geometric measures.

\subsection{Geometric measure theory applied to surfaces}
Let $S$ be a parameterized surface embedded in $\mathbb{R}^3$. This surface and its triangulations can be understood as points in a high-dimensional space of shapes (infinite in the case of continuous surfaces). We look for a robust metric on this particular space, suitable for unsupervised learning tasks. If we take a diffeomorphism $\phi$ acting on the parameter space, such a metric should not differentiate between a given shape $S$ and a reparameterization $\Tilde{S} = S \circ \phi$ of this shape: $S$ and $\Tilde{S}$ are at distance 0 for this metric.

\begin{definition}[Varifold representation of surfaces]

The varifold $\mu_S$ associated to  a continuous surface shape $S$ is the measure on $\mathbb{R}^3 \times \mathbb{S}^2$
such that, for any continuous test function $u: \mathbb{R}^3 \times \mathbb{S}^2 \mapsto \mathbb{R}$:
\begin{equation}
    \mu_S (u) = \int_{\mathbb{R}^3\times \mathbb{S}^2} ud\mu_S = \int_S u(x, {\vec{n}_x})\text{d}\sigma(x)
\end{equation}
where $\vec{n}_x$ is the normal of $S$ at $x$ and $d\sigma$ the area measure of the surface $S$.
\end{definition}

The key property that motivates the use of varifolds in our context is that for any two parameterized shapes $S$ and $\Tilde{S}$, $\mu_S=\mu_{\Tilde{S}}$ if and only if $\Tilde{S}$ is a reparameterization of $S$ \cite{Charon_varifolds}. 

Moreover, there is a natural discrete version of varifolds for meshes as follows. If $f$ is a triangle (e.g. a face in a triangular mesh) with center $c(f)$, and normal $\vec{n}_f$, the corresponding discrete varifold $\mu_f$ is given by a Dirac mass $\delta_{c(f)}^{\vec{n}_f}$ at $(c(f),\vec{n}_f)$ weighted by $a(f)$ the area of $f$. In other words, for any continuous test function $u$ on $\mathbb{R}^3\times\mathbb{S}^2$, 
$$
\mu_f(u):=a(f)u(c(f),\vec{n}_f).
$$
and we can write $\mu_f=a(f)\delta_{c(f)}^{\vec{n}_f}.$ Therefore, we can extend this to a triangular mesh:

\begin{definition}[Discrete varifold representation of surfaces]

Let a mesh $X = \{V(X), E(X), F(X) \}$, where $V(X)$, $E(X)$, and $F(X)$ are respectively, the set of vertices, edges, and faces. The varifold representation associated to $X$ is the measure on $\mathbb{R}^3 \times \mathbb{S}^2$, given by 
$$
\mu_{X}:=\sum_{f\in F(X)}\mu_f=\sum_{f\in F(X)}a(f)\delta_{c(f)}^{\vec{n}_f},
$$
with $\mu_f=a(f)\delta_{c(f)}^{\vec{n}_f}$ as described above. 
\end{definition}
This representation is well suited for triangular meshes as each triangle is represented by a measure on the position of its center $c(f)$ and its orientation $\vec{n}_f$ given by a point on the 2-sphere, all weighted by the area of the triangle $a(f)$. 

Moreover, we have the following result, which is an easy consequence of Proposition 1 from \cite{kaltenmark2017general} combined with Corollary 1 from \cite{MORVAN_closely_inscribed}:
\noindent
\begin{theorem}\label{th:approx}
    Take $u:\mathbb{R}^3\times\mathbb{S}^2\rightarrow \mathbb{R}$ a bounded $k_u$-Lipschitz function, with supremum $\Vert u\Vert_\infty$. Let $S$ a be surface, and $X$ a triangular mesh drawn from $S$ whose vertices belongs to $S$, with greatest edge length $\eta_X$, and smallest angle $\theta_X$ among its faces. Denote $\kappa_S$ the greatest principal curvature over $S$, and $a(S)$ its surface area. Then, there is a universal constant $C$ such that
    $$
    \vert \mu_S(u)-\mu_X(u)\vert\leq Ca(S)\frac{(\kappa_S+1)}{\sin\theta_X}(k_u+\Vert u\Vert_\infty)\eta_X
    $$
\end{theorem}
Consequently, for any two good triangulations with relatively small edges $X$ and $\hat{X}$ of $S$, $\mu_X \simeq \mu_{\hat{X}}$, making discrete varifolds natural tools for mesh-invariant purposes.

\noindent\textbf{Comparing shapes with kernel metrics.}

\noindent With this representation of shapes, we compute dissimilarities between shapes, both continuous and discrete, by using kernels on $\mathbb{R}^3\times\mathbb{S}^2$. Following the works of ~\cite{Vaillant_Glaunes_current_2005,Charon_varifolds, kaltenmark2017general,PiersonPierson2022}, we use a product $k=k_pk_n$, with $k_p$ a kernel on $\mathbb{R}^3$ and $k_n$ a kernel on $\mathbb{S}^2$.

To ensure invariance under the action of rigid motions (rotations and translations), we choose a radial basis function $\rho$ to drive the position kernel $k_p$ and a zonal kernel $\gamma$ for the orientation kernel $k_n$. Details on admissible functions for $\gamma$ and $\rho$ can be found in \cite{positive_kernels, metric_space_kernels}.
\begin{align}
    k_p : \begin{cases}
        \mathbb{R}^3 \times \mathbb{R}^3 &\rightarrow \mathbb{R}\\
        (x, \hat{x}) &\mapsto \rho(|x - \hat{x}|)
    \end{cases}
\end{align}
\begin{align}
    k_n:\begin{cases}
        \mathbb{S}^2 \times \mathbb{S}^2 &\rightarrow \mathbb{R}\\
        (\vec{n}_x, \vec{n}_{\hat{x}}) &\mapsto \gamma(\langle \vec{n}_x, \vec{n}_{\hat{x}} \rangle)
    \end{cases}
\end{align}

These kernels are extrinsic in the sense that they are defined on the ambient space $\mathbb{R}^3 \times\mathbb{S}^2$, and use the euclidean distances. 

Then, we can derive a \noindent correlation between any two measures $\mu,\hat{\mu}$ on $\mathbb{R}^3\times\mathbb{S}^2$ as
$$
\left<\mu,\hat{\mu}\right>_k=\int_{\mathbb{R}^3\times\mathbb{S}^2}\int_{\mathbb{R}^3\times\mathbb{S}^2}
k_p(x,\hat{x})
k_n(\vec{n}_x,\vec{n}_{\hat{x}})d\mu(x,\vec{n})d\hat{\mu}(\hat{x},\vec{n}_{\hat{x}})
$$

\noindent This gives a parameterization-independent correlation between two surfaces $S$ and $\hat{S}$ through the kernel $k$ as follows:
\begin{equation}
\label{eq:varifold_dot_continuous}
    \langle \mu_S, \mu_{\hat{S}} \rangle_k = \iint_{S\times \hat{S}}k_p(x,\hat{x})k_n(\vec{n}_x, \vec{n}_{\hat{x}})\text{d}\sigma(\hat{x})\text{d}\sigma(x)
\end{equation}


\noindent For the discrete setting, we write the correlation between two faces $f$ and $\hat{f}$ 
    \begin{equation}
        \langle f, \hat{f} \rangle_k = a(f) a(\hat{f}) k_n(\vec{n}_{f}, \vec{n}_{\hat{f}}) k_p(c(f), c(\hat{f})).
    \end{equation}
    
\noindent This can be summed along the meshes to give a discretized version of ~\Cref{eq:varifold_dot_continuous}. For $X$ and $\hat{X}$ two meshes, the correlation is 
\begin{equation}
    \langle \mu_X, \mu_{\hat{X}} \rangle_k = \sum_{f\in F(X)}\sum_{\hat f\in F(\hat{X})} a(f) a(\hat{f}) k_p(c(f),c(\hat{f}))k_n(\vec{n}_{f},\vec{n}_{\hat{f}})
\end{equation}
    
\noindent Now for some kernels, these formulas actually give a positive definite dot-product on the space of measures, so that $\mu=\hat{\mu}$ if and only if $\Vert \mu-\hat{\mu}\Vert^2=\left<\mu-\hat{\mu},\mu-\hat{\mu}\right>=0$. From there, we define the "geometric measure" (GM) loss associated to such a kernel by
\begin{align}
    \mathcal{L}_k^{GM}(X, \hat{X}) &= \langle \mu_{X}-\mu_{\hat{X}}, \mu_{X}-\mu_{\hat{X}} \rangle_{k} \nonumber \\&= \langle \mu_{X}, \mu_{X} \rangle_{k} + \langle \mu_{\hat{X}}, \mu_{\hat{X}} \rangle_{k}
    - 2\langle \mu_{X}, \mu_{\hat{X}} \rangle_{k}.
\end{align}
For well-chosen kernels, this function is fully differentiable and can be written in closed form, making this loss function suitable for GPU accelerated computations. Moreover, thanks to Theorem \ref{th:approx}, this GM loss is robust to a mesh change in both $X$ and $\hat{X}.$

\begin{figure}[h!]
    \centering
    \includegraphics[width=0.95\linewidth]{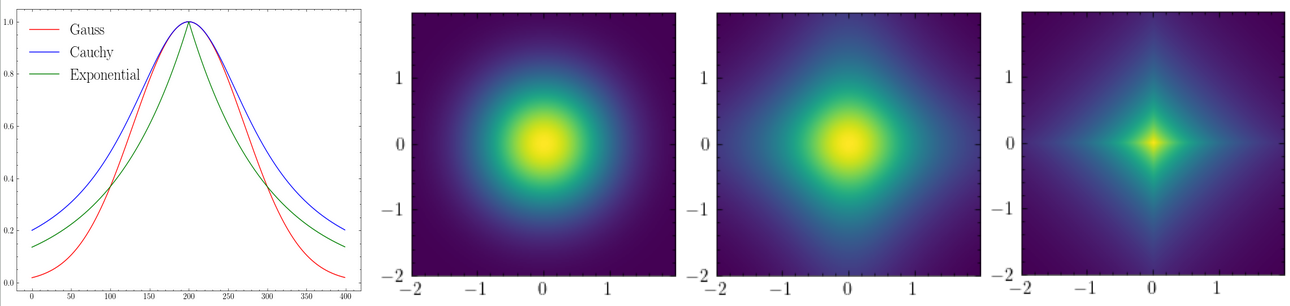}
    \caption{Visualization of 3 kernels suitable for $k_p$, namely a Gaussian kernel of the form $x \mapsto \exp \left( \frac{x^2}{\sigma^2}\right)$, a Cauchy kernel of the form $x \mapsto \frac{1}{1 + \left(\frac{x}{\sigma}\right)^2}$ and an exponential kernel $x \mapsto \exp\left(\frac{|x|}{\sigma}\right)$. On the right is displayed the 1D plot of the three kernels: in red the Gaussian kernel, in blue the Cauchy kernel and in green the exponential one. Next,} 2D plots of the Gaussian, Cauchy and exponential kernel respectively from left to right.
    \label{fig:viz_kernels}
\end{figure}

\textbf{Choice of kernel and loss function.} Several kernels are suitable for $k_p$ such as Gaussian, linear and Cauchy kernels. We display some of them in~\Cref{fig:viz_kernels}.

By default, we use a Gaussian kernel for the position and, depending on the type of geometric measure, we either use a linear (current), squared (varifold) or exponential (oriented varifold) zonal kernel on $\mathbb{S}^2$ for $k_n$. In particular, $k_p$ is defined by a scale parameter $\sigma$. Typically, this parameter should be chosen in order to encompass the structure of local neighborhoods across the meshes.


To improve the versatility and efficiency of our loss, we propose to use a sum of kernels with different scale parameter for each term, such that, our final loss is defined as
\begin{equation}
    \mathcal{L}^{GM} = \sum_i \lambda_i \mathcal{L}_{k_i}^{GM}
\end{equation}
with $k_i$ associated with the scale $\sigma_i$ and a scalar weighting coefficient $\lambda_i$. The number of kernels and the coefficients $(\lambda_i)_i$ are hyperparameters of the model. For our task, we observed experimentally that setting $\lambda_i = \left(\frac{\sigma_i}{\max_i \sigma_i}\right)^2$ could gives good enough results. This way, we penalize small scales but still allow the metric to distinguish fine structures on the mesh made up of small triangles.

\subsection{Application: face generation}
Human face modeling involves deformable geometries with small but meaningful changes. In general, we model a face shape $S$ as the combination of an identity (shape of the face) and an expression (reversible deformation from a neutral expression): this is the so-called \textbf{morphable model} \cite{Egger_3DMorphableSOTA_2020}, \cite{Daoudi3DfaceBook}. It has seen many improvements over the past few years thanks, in part, to computer vision research and deep learning. Indeed, using nonlinear, deep representations presents the potential to outperform traditional linear or multi-linear models in terms of generalization, compactness, and specificity. In addition, the implementation of deep networks for parameter estimation allows for quick and dependable performance even with uncontrolled data.
\begin{align}
    S = \text{Id} + \text{Expr}= \text{Template} + \Delta \text{Id} + \Delta \text{Expr}
\end{align}
and we set $\Delta = \Delta \text{Id} + \Delta \text{Expr}$ the total deformation of the template to match the target face.

\begin{figure}[h!]
    \centering
    \includegraphics[width=.9\linewidth]{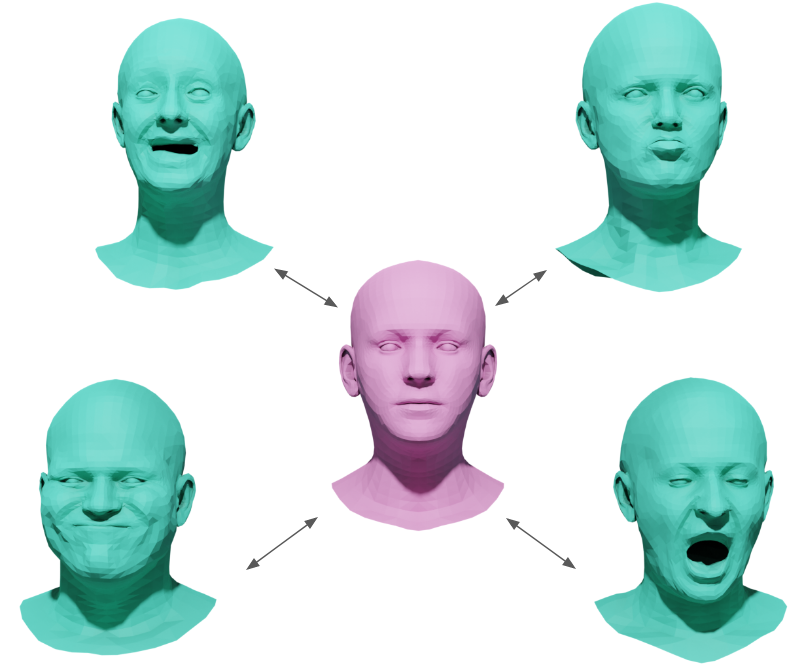}
    \caption{Illustration of a morphable model: a template mesh (in red) is deformed into a given mesh (in blue).}
    \label{fig:morphable_model}
\end{figure}

\noindent The COMA dataset \cite{COMA} encapsulates this representation as it is made of 12 identities executing 12 different expressions. Each expression is a sequence of meshes during which the subject starts from a neutral face, executes the expression and goes back to a neutral face. Each sequence is made from 25 to more than 200 meshes and the sequence length is not consistent across the identities. A distinction is made between the unregistered meshes obtained from scans and their registered counterpart providing us with two distinct databases.

\section{Experiments}
The model corresponding to the architecture presented in~\Cref{fig:autoencoder} is trained end-to-end, both the encoder and the decoder weights are optimized at the same time.
The Python code is built over the one from ~\cite{bouritsas_neural3DMM_2019} using the Pytorch framework. All measurements were conducted using the same machine (a laptop) with an NVIDIA Corporation / Mesa Intel® UHD Graphics (TGL GT1) as GPU and Intel® Core i5-9300HF 2.40GHz CPU with 8,00 Go RAM.

\subsection{Implementation details}

Our model takes as input a mesh of any parameterization and gives out a mesh in the COMA topology which has 5023 vertices and 9976 faces with a fixed graph structure (the topology of the output solely depends on the topology of the chosen template which can be adapted).\\

The encoder is a combination of a simple PointNet architecture (PN), a spatial transformer (TF) as described in \cite{spatial_transformer} to improve invariance to euclidean transformations of the input and a fully connected layer (FC). We use the parameters of \cite{cosmo2020limp}, that were optimized on the COMA dataset. The filter sizes for the decoder are [128, 64, 64, 64, 3]. It starts with a fully connected layer (FC) and then alternately performs up-sampling (US) and spiral (de)-convolution (SC). The parameters are taken from the original Neural3DMM \cite{bouritsas_neural3DMM_2019} paper.

\begin{itemize}
    \item \textit{Encoder: } PN(64,1024) $\rightarrow$ TF(64) $\rightarrow$ FC(128,64,64)
     \item \textit{Decoder: } FC(128) $\rightarrow$ US(4) $\rightarrow$ SC(128) $\rightarrow$ US(4) $\rightarrow$ SC(64) $\rightarrow$ US(4) $\rightarrow$ SC(64) $\rightarrow$ US(4) $\rightarrow$ SC(64)
\end{itemize}

The model learns to code the deformation from a fixed template mesh $X^t$ that does not belong to the training
dataset. It is trained for 100 epochs with a latent space of fixed size 128. We used Adam optimizer with a learning rate of $10^{-3}$ and a batch size equal to 16.

\begin{figure*}[ht]
    \centering
    \includegraphics[width=.9\linewidth]{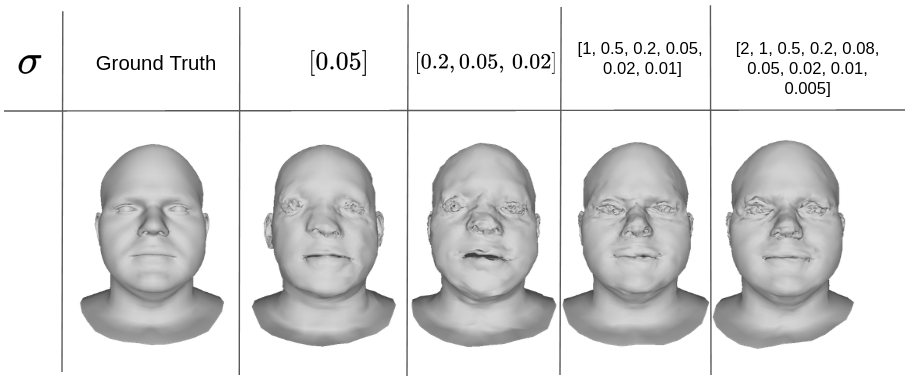}
    \caption{Qualitative comparison for the reconstruction of a target mesh (on the left) with a growing number of kernel (summed) as we go on the right.}
    \label{fig:comparison_multi_scale}
\end{figure*}

\noindent Unfortunately, this loss alone is optimal when the size of the triangles is relatively constant along the mesh. In order to cope with this limitation, we use what we call a \textbf{multi-kernel metric} with a collection of different scale $(\sigma_i)_i$. Experimentally, we observe that the loss function induces poor performances when we do not use enough kernels but also when we use to much of them (see~\Cref{fig:comparison_multi_scale}). 

Even with a carefully optimized loss function, we still observe some noise in the reconstruction. To correct this, we propose to use a post-processing step with a Taubin smoothing \cite{Taubin_smoothing_1995}. We highlight the improvement in quality of the reconstructed mesh in~\Cref{fig:viz_reconstruction}. More sophisticated techniques can be applied such as using the pretrained model from Kim \etal \cite{Kim_Roh_Im_Kim_SpiralDenoise_2022} to remove the noise.

\subsection{Learning performances}
We train our model on 11 out of the 12 identities of the COMA dataset and we evaluate the performance on the remaining identity. Therefore, we assess the generalizability of the model and we compare the performances to other unsupervised methods. FLAME \cite{FLAME_2017} is a face morphable model. In the paper, the authors use this model to register faces easily. They minimize a regularized Chamfer distance and use landmark information to register a face model. To be fair to our method, we do not use landmark information and minimize the mesh-to-mesh distance. 3DCODED \cite{groueix_3DCODED_2018} is a deep learning model using a Pointnet encoder and a MLP decoder that deforms a template mesh. Their unsupervised version uses Chamfer distance as a loss function, and is regularized using an edge loss and a Laplacian loss. This model has demonstrated its efficiency for human body registration, and we modify it in order to apply it on human faces. We also tried a more recent model: Deep Diffeomorphic Registration \cite{Croquet_Diff_Reg_OT_2021} (DDFR). It is a deep learning model that learns a diffeomorphism of the ambient space in order to morph a source mesh (the template) to a target mesh. Unfortunately, the training process has shown overflowing computational cost while showing limiting ability to produce expressive faces. This can be explained as diffeomorphisms of the ambient space can hardly separate the lips and produce "sliding motions" which are not differentiable.

We compute the reconstruction error according to 3 metrics: the Hausdorff distance $d_H$, the Chamfer distance $d_{CD}$ and a Varifold distance $d_V$ with $k_p$ being a Gaussian kernel with $\sigma=0.1$. This specific scale has been chosen as it is roughly ten times the average size of a triangle. The results are presented in~\Cref{tab:perfs_comparison}.

\begin{table}[ht!]
\centering
\begin{tabular}{l|l|l|l}
\hline
Model   & $d_H$ & $d_{CD}$ ($\times 10^{-4}$) & $d_V$ ($\times 10^{-4}$) \\ \hline
3DCODED \cite{groueix_3DCODED_2018} & 0.018 & 0.312 & 0.267 \\
FLAME \cite{FLAME_2017} & 0.012 & 0.109 & 0.013 \\
\hline
Ours  & 0.010 & 0.091 & 0.011 \\
Ours (+ filter)   & \textbf{0.009} & \textbf{0.088} & \textbf{0.011}   
\end{tabular}
\caption{Reconstruction error when learning faces from 11 out of the 12 identities of COMA and tested on the remaining identity which constitutes around 1200 meshes for the test set. The error is averaged along this test set.}
\label{tab:perfs_comparison}
\end{table}

\begin{figure*}[h!]
    \centering
    \includegraphics[width=.98\linewidth]{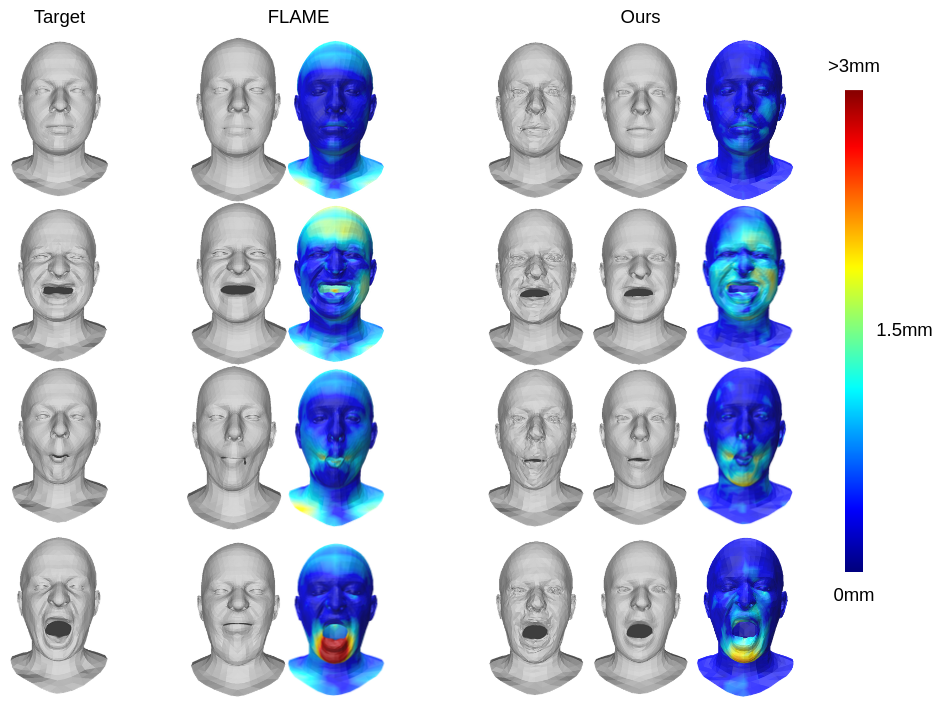}
    \caption{Qualitative results on the reconstruction of facial expressions from the COMA dataset with our AE model. On the left is the target mesh, the first pair of reconstructions shows the reconstructed mesh with a linear method (FLAME) and on the right is displayed the reconstruction with our model. The first proposed reconstruction is obtained without any post-processing and the other one with a Taubin smoothing (parameters: $\lambda=0.5$, $\mu=-0.53$) with its corresponding MSE heatmap of error.}
    \label{fig:viz_reconstruction}
\end{figure*}

We highlight the following observations:
\begin{enumerate}
    \item The FLAME model struggles to reconstruct "extreme" facial expressions such as a wide opened mouth in the 4th row of figure~\Cref{fig:viz_reconstruction}. We also report a longer time for the registration.
    \item 3DCODED, while being a lot faster than linear methods such as FLAME, shows poor performance for learning human faces. In fact, the model only output indistinguishable deformations from the template mesh.
    \item Our model surpass FLAME in term of expressivity and 3DCODED in terms of efficiency.
\end{enumerate}

\subsection{Ablation study}

Next, we show that the effectiveness of our model stems in particular from the loss function. We compare some unsupervised losses mentioned in section 3 with our geometric measure based loss $\mathcal{L}^{GM}$ without changing any other parameters. We focus on the encoding and decoding of 4 sequences of expression ($\#1$: \textit{bareteeth}, $\#2$: \textit{cheeks\_in}, $\#3$: \textit{high\_smile} and $\#4$: \textit{mouth\_extreme} in~\Cref{tab:losses_comparison}).
Here, $\mathcal{L}^{GM}$ is computed with a Gaussian kernel for $k_p$ and a squared zonal kernel for $k_n$ which correspond to the framework of unoriented varifolds.
\begin{table*}[t!]
\centering
\scriptsize
\setlength\tabcolsep{2pt} 
\begin{tabular}{c}
    \begin{adjustbox}{max width=\textwidth}
    \aboverulesep=0ex
    \belowrulesep=0ex
    \renewcommand{\arraystretch}{1.0}
    \begin{tabular}[t]{l|llll|llll|llll}
Loss function   & \multicolumn{4}{c|}{Hausdorff}    & \multicolumn{4}{c|}{Chamfer ($\times 10^{-4}$)}     & \multicolumn{4}{c}{Varifold ($\times 10^{-4}$)}      \\ \hline
                & \#1 & \#2 & \#3 & \#4 & \#1 & \#2 & \#3 & \#4 & \#1 & \#2 & \#3 & \#4 \\ \cline{2-13} 
$\mathcal{L}^{DCD}$      & 0.031  & 0.031  & 0.030  & 0.031  &  0.45 &  0.44 & 0.42  & 0.44  & 0.38 & 0.38 & 0.38  & 0.38  \\
$\mathcal{L}^{DCD}+ \mathcal{L}^{edges}$   & 0.030  & 0.029  & 0.030  & 0.029  &  0.43 &  0.42 & 0.43  & 0.45  & 0.35 & 0.34 & 0.36 & 0.34 \\
$\mathcal{L}^{SD} (\epsilon=0.0001)$    & 0.018 & 0.017 & 0.017 & 0.017  & 0.101 & 0.099  & 0.094  & 0.102 & 0.016 & 0.017 & 0.015 & 0.016 \\
$\mathcal{L}^{SD} (\epsilon=0.001)$    & 0.024  & 0.027 & 0.023 & 0.023  & 0.135  & 0.133   &  0.120  &   0.134   & 0.023 & 0.023 & 0.021 & 0.022  \\
$\mathcal{L}^{SD} (\epsilon=0.01)$    & 0.036 & 0.041 &  0.030 & 0.031  &  0.218 &  0.198   &  0.152  &  0.155  & 0.030 & 0.035 & 0.032 & 0.034  \\
$\mathcal{L}^{GM}$ (ours)   &   0.010  &  0.011  &  0.09 &  0.010  & 0.089  &  0.084   &  0.086   &  0.085     & 0.011 & 0.011 & 0.010  &  0.011    
    \end{tabular} 
    \end{adjustbox}
          
    \end{tabular}
\caption{Ablation study: We trained our model using different unsupervised losses (only the loss is changed) and we report the mean reconstruction error when learning expressions in the COMA dataset. The reported evaluation is performed on meshes that does not belongs to the training data. We account for the quality of the learning process in relation to the loss function employed for the task.}
\label{tab:losses_comparison}
\end{table*}

We conducted the test using $\mathcal{L}^{SD}$ with different value of approximation $\epsilon$ as it is the most challenging candidate to surpass $\mathcal{L}^{GM}$. In spite of this, setting $\epsilon$ to be less than $10^{-4}$ showed declining performances in addition to much higher computational cost.

\subsection{Robustness}
\begin{figure}[h!]
    \centering
    \includegraphics[width=.9\linewidth]{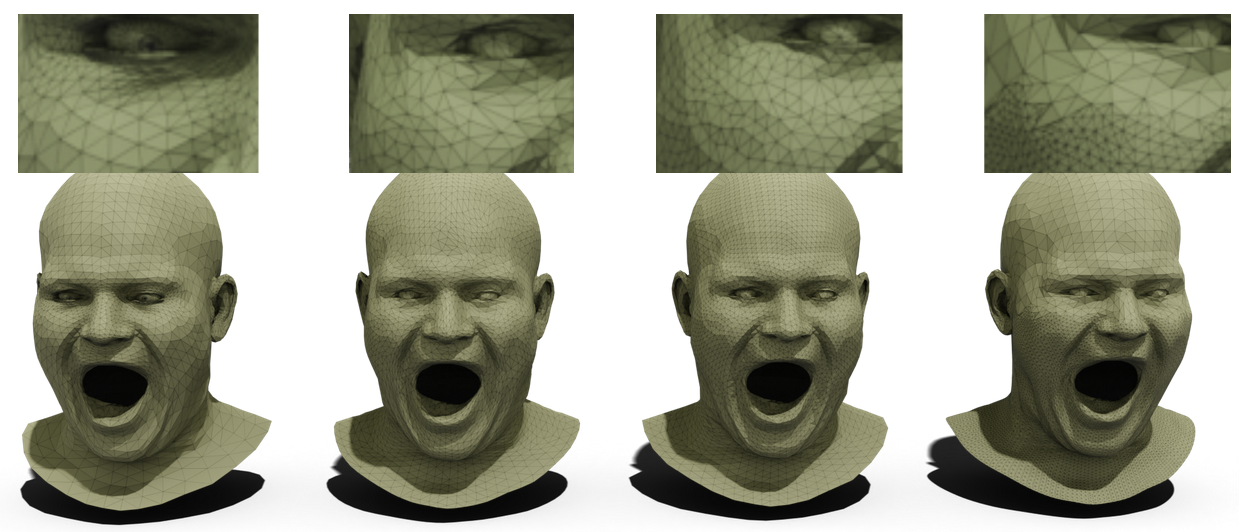}
    \caption{Examples of reparameterized meshes of a single expressive face, from left to right: original parameterization, \textit{UpDown}, \textit{Iso} and \textit{Variable}}
    \label{fig:viz_reparams}
\end{figure}

We evaluate our model on different reparameterised meshes of faces to demonstrate that our model learns the geometry of the shape instead of the graph structure. This experience is conducted on one identity executing all 12 expressions. The proposed reparameterizations of the meshes are displayed in~\Cref{fig:viz_reparams}.

In a similar fashion as in \cite{sharp2022diffusionnet}, we test the robustness of our model against three different reparameterizations. \textit{UpDown} is obtained by subdividing the mesh and then performing a quadric edge simplification. \textit{Iso} is the result of one iteration of explicit isotropic remeshing. Finally, the \textit{Variable} parameterization is obtained by dividing the mesh in two-half: the top and the bottom part. On the top part, we perform a simplification to diminish the number of triangles and on the bottom part, we subdivide the mesh in order to get a much higher density of triangles. These new meshes are obtained via Meshlab \cite{Meshlab} remeshing routines, using the \textit{pymeshlab}\footnote{\url{https://pymeshlab.readthedocs.io/}} Python library.

\begin{figure}[ht!]
    \centering
    \includegraphics[width=.95\linewidth]{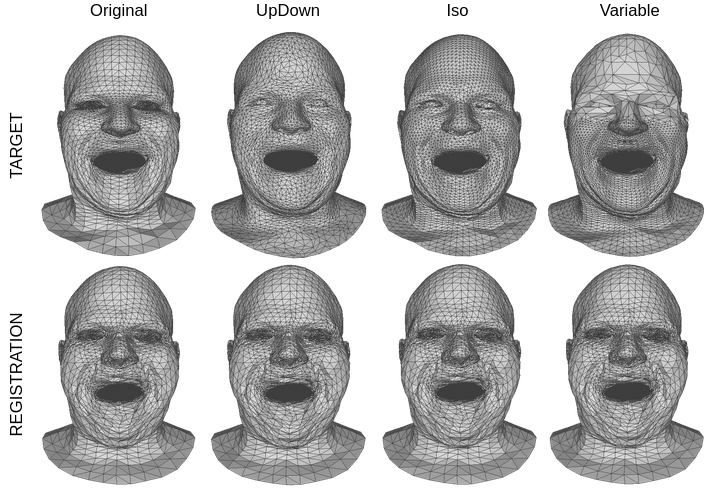}
    \caption{An example of a mesh, its three reparameterizations and the corresponding registration with our model.}
    \label{fig:robustness_ex}
\end{figure}

We stress the robustness of our model by comparing the outputs of the model when given with the original parameterisation as input and the reparameterization as input. The results are summarized in~\Cref{tab:perfs_robustness}, where we display the relative difference between the two outputs. 
\begin{table}[h!]
\centering
\begin{tabular}{l|l|l}
\hline
   & Hausdorff & Chamfer \\ \hline
Original - Original &    0      &   0    \\
Original - UpDown   &  0.0035      &   0.0096   \\
Original - Iso      &  0.0012      &   0.0016   \\
Original - Variable &  0.0018      &   0.0030
\end{tabular}
\caption{Relative difference of reconstruction when the model is tested on reparameterized meshes. The evaluation is performed on  meshes that do not belong to the training set in order to evaluate the robustness of the learning process.}
\label{tab:perfs_robustness}
\end{table}
While we have indeed a slight difference between the outputs, the worst relative difference is around $2 \%$ in Chamfer distance. Qualitative results that we display in~\Cref{fig:robustness_ex}, highlight this robustness visually.

\subsection{Training on inconsistent batches of meshes}
\begin{figure*}[h!]
    \centering
    \includegraphics[width=.98\linewidth]{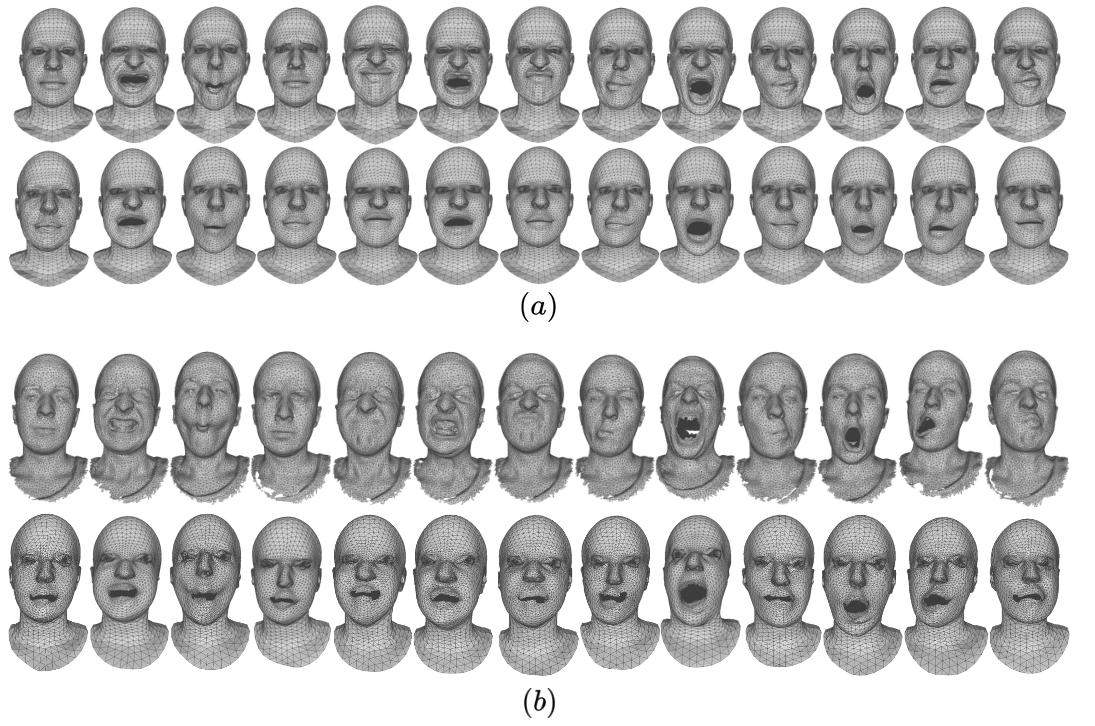}
    \caption{(a): Registered case, fixed resolution, point-wise correspondence and no noise. (b): Raw scans, variable resolution, no correspondence and noise.
}
    \label{fig:test_raw_scans}
\end{figure*}

As stated above, our model is able to train on mesh data with variable resolution, hence it can be trained on sampled scans from COMA which makes an inconsistent database. We show that the model is still capable of learning identity and expressions. The only required pre-processing step is a rigid alignment (with scaling) to the chosen template. \\
As the scans are highly detailed (several gigabytes for each subject), we train our model on one identity with its 12 expressions. We compare the results of our model trained on the registered meshes against another model, with same parameters, trained on the raw scans. We synthesize the results in~\Cref{tab:comparison_train_registered_raw} and display a few examples of reconstruction in~\Cref{fig:test_raw_scans}.

\begin{table}[]
\centering
\begin{tabular}{lcccc}
\hline
Expression     & \multicolumn{2}{c}{Registered} & \multicolumn{2}{c}{Raw scans} \\ \hline
               & Hausdorff       & Chamfer      & Hausdorff      & Chamfer      \\ \hline
Neutral        &     0.009            &   0.075          &    0.014            &   0.24           \\
bareteeth      &     0.010            &   0.075         &    0.014            &   0.26           \\
cheeks\_in     &     0.010            &   0.074          &    0.014            &   0.25           \\
eyebrow        &     0.009            &   0.076          &    0.014            &   0.28           \\
high\_smile    &     0.010            &   0.076          &    0.014            &   0.26           \\
lips\_back     &     0.009            &   0.073          &    0.014            &   0.25           \\
lips\_up       &     0.009            &   0.075          &    0.015            &   0.24           \\
mouth\_down    &     0.009            &   0.077          &    0.014            &   0.25           \\
mouth\_extr &     0.009            &    0.075          &    0.012            &   0.24           \\
mouth\_mid  &     0.009            &    0.074          &    0.013            &   0.27           \\
mouth\_open    &     0.009            &   0.074          &    0.013            &   0.28           \\
mouth\_side    &     0.009            &   0.076          &    0.013            &   0.27           \\
mouth\_up      &     0.009            &   0.075          &    0.014            &   0.29          
\end{tabular}
\caption{Reconstruction error for each expression}
\label{tab:comparison_train_registered_raw}
\end{table}

\subsection{Evaluating the latent space}

The advantage with our solution is that we can directly operate in the latent space to deform any face. We use this property for three applications, interpolation between faces, extrapolation of a face motion and expression transfer between faces.

\begin{figure}[h!]
    \centering
    \includegraphics[width=.95\linewidth]{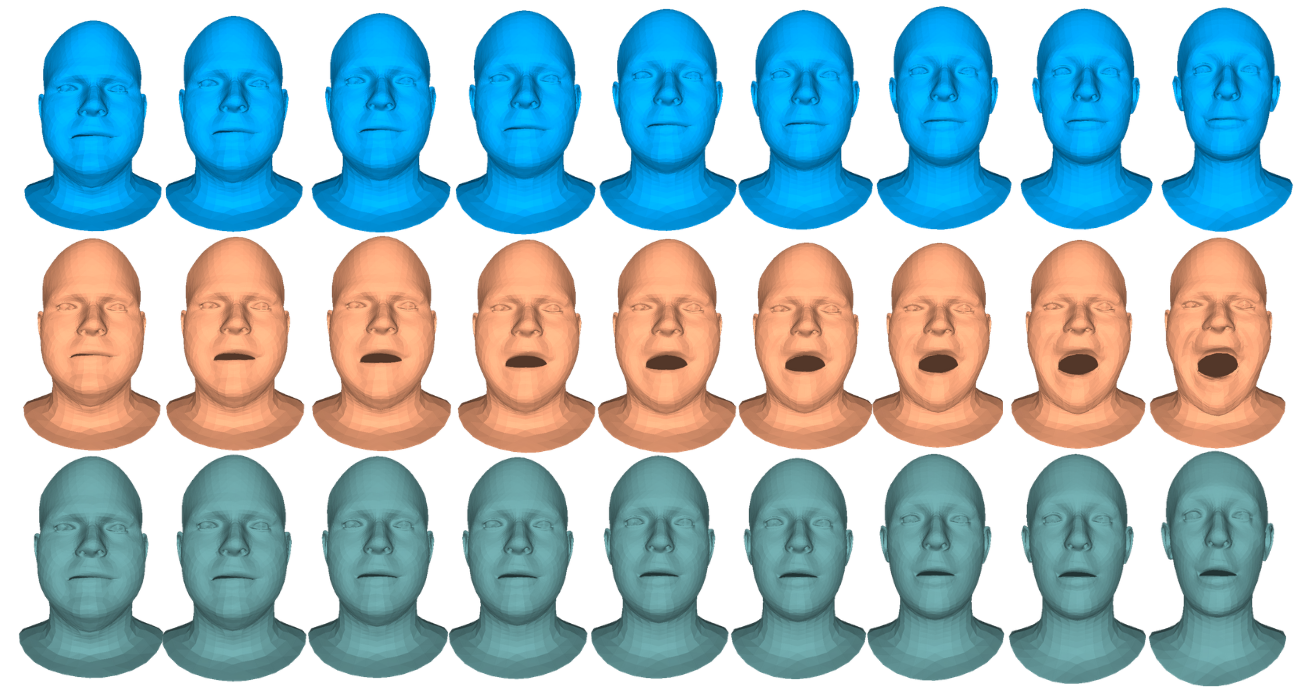}
    \caption{Linear interpolation in the latent space. The first row shows an interpolation between two neutral identities, the second row shows the result of an interpolation between a neutral face and one of its expressions. The last row shows an interpolation between a neutral identity and an expressive other identity.}
    \label{fig:interpolation_latent}
\end{figure}

\noindent \textbf{Interpolation. }We compute a linear interpolation $(z_t)_{t \in [0,1]}$ between a source latent vector $z_0$ and a target one $z_1$ with
$$
    z_t = (1 - t)z_0 + t z_1.
$$
We display the resulting interpolation on faces in~\Cref{fig:interpolation_latent}, between two identities, two poses and the interpolation of two faces with both characteristics being different. The figure show that the results are visually satisfying.

\begin{figure}[h!]
    \centering
    \includegraphics[width=.95\linewidth]{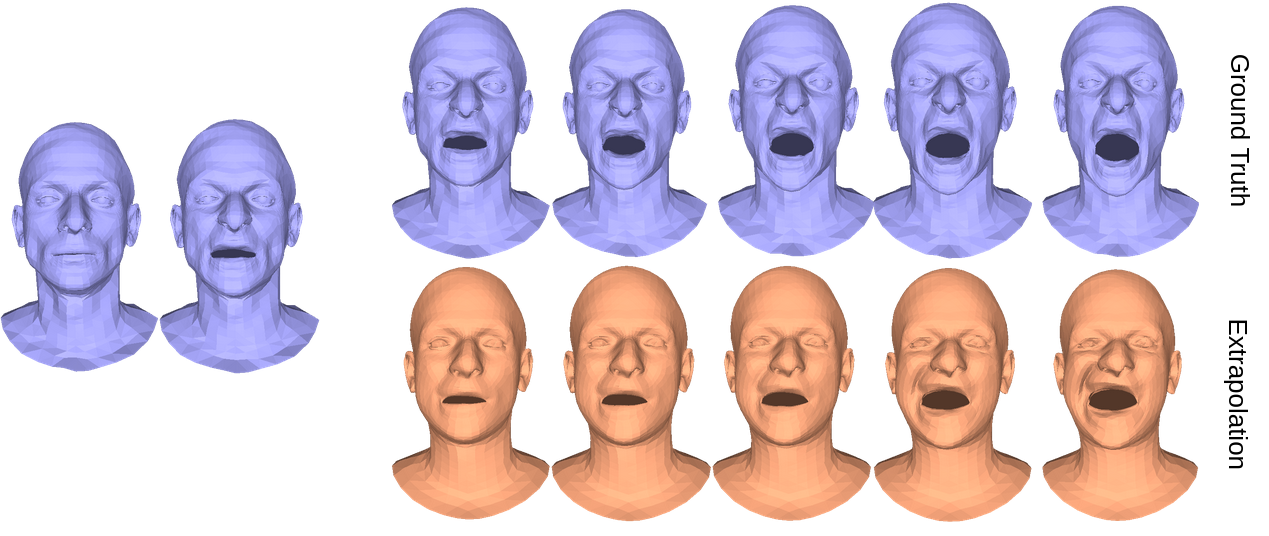}
    \caption{Linear extrapolation (in orange) from two meshes (on the left) of a sequence describing an expression.}
    \label{fig:extrapolation_latent}
\end{figure}

\noindent \textbf{Extrapolation.} Given a initial motion of a face (two close meshes starting a motion), we would like to extrapolate the full motion. This can be formulated easily in the latent space: from the two meshes' latent codes $z_1, z_2$, we shoot a time dependent path $z_t$ from the inital speed $(z_2 - z_1)$:
$$
    z_t = z_1 + t (z_2 - z_1).
$$
We display the resulting motions in~\Cref{fig:extrapolation_latent}. We observe that the desired motion is reproduced. At large time steps, some non natural deformations start to appear, but we are still able to recognize the expression of the face.

\begin{figure}[h!]
    \centering
    \includegraphics[width=.95\linewidth]{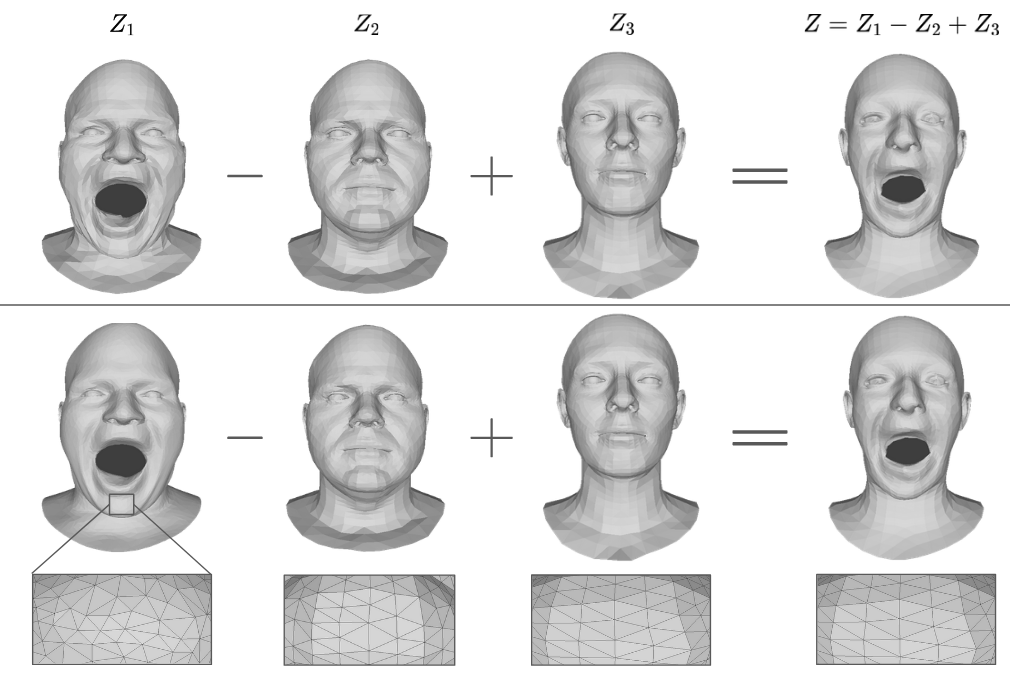}
    \caption{An example of a robust expression transfer: from an expressive face encoded as $Z_1$, we subtract its neutral identity encoded as $Z_2$ and replace it with another one encoded as $Z_3$.
}
    \label{fig:robust_arithmetic_latent}
\end{figure}

\noindent  \textbf{Expression transfer in the latent space. }Thank to the auto-encoding architecture, we also demonstrate its ability to perform complex mesh manipulation such as expression transfer with simple arithmetic operations in the latent space (additions and subtractions). We also demonstrate the robustness of such operations as in~\Cref{fig:robust_arithmetic_latent}.

\begin{figure}[h!]
    \centering
    \includegraphics[width=.95\linewidth]{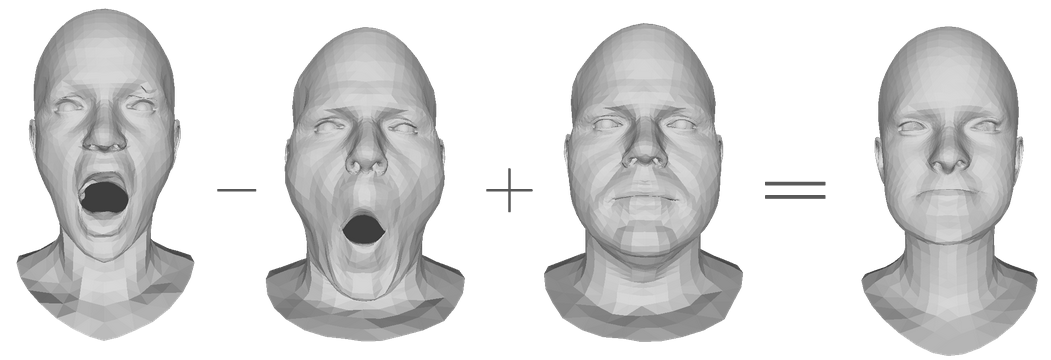}
    \caption{An example of neutralization of an expression to recover a neutral face. From an expressive face on the left, we subtract a similar expression from another identity, which we cancel with its corresponding neutral face to obtain an estimation of the neutral face of the first identity (on the right).
}
    \label{fig:expr_neutralization}
\end{figure}

\noindent In a similar fashion, manipulating the latent space to cancel an expression and recover the neutral face is possible as shown in~\Cref{fig:expr_neutralization}.

\section{Discussion}
In this last section, we discuss some areas for improvement and implications of the presented work.

\subsection{Complexity}
Regarding the complexity of our model, it is made of simple elements and most of the computation complexity comes from the calculation of the loss at each epoch, during training.\\
The computation of the kernel metric is already optimised with the KeOps library \cite{Charlier_Feydy_KeOps_2021} which uses symbolic matrices to avoid memory overflow. Consequently, training has a squared polynomial complexity.\\
Once the model is trained, the registration is performed with a simple function evaluation which makes it a lot faster than non-learning method such as FLAME, elastic matching or LDDMMs.\\
We report the training time for a single epoch, having the setting detailed in 4.1, in~\Cref{tab:time_per_epoch}.

\begin{table}[h!]
\centering
\begin{tabular}{llll}
Model & \multicolumn{1}{c}{3DCODED} & Ours ($\mathcal{L}^{DCD}$) & \multicolumn{1}{c}{Ours ($\mathcal{L}^{GM}$)} \\ \hline
Time  &     2min38    &     3min30   &     17min                     
\end{tabular}
\caption{Mean time per epoch reported during our experiment}
\label{tab:time_per_epoch}
\end{table}

\subsection{Limitations}
As we can observe, some rexgions with high curvature are hardly reconstructed, especially around the eyes, the lips and the nostrils. We believe it is due to the fact that the varifold struggles to take into account both large and finer structures on the mesh. Therefore, it is possible that the model can be enhanced using normal cycles \cite{Roussillon_Glaunes_2016} which take curvature information into account. But this comes at a high computational cost.

\begin{figure}[h!]
    \centering
    \includegraphics[width=.5\linewidth]{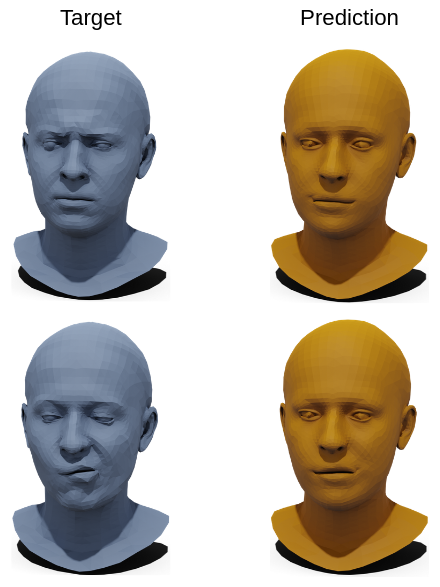}
    \caption{Examples of failed reconstruction of the expression. On the left is displayed the target and on the right its reconstruction with our model. The identity is preserved but the model fails to encode the expression.}
    \label{fig:limits}
\end{figure}

We also point out that our model is limited by the encoder which certainly has the advantage of being robust. But, this robustness comes at a cost regarding the performances as we show in~\Cref{fig:limits}. Indeed, the encoder of our model struggles to keep small deformations (such as a slight eyebrow movement) during the encoding part.

Overall, we believe that using our loss function based on geometric measures has the potential to yield superior results than the Chamfer distance which is widely used for similar tasks. Indeed, during our experiments, the Chamfer loss has shown poor capability to set an effective objective function, especially when used to generate faces.

\subsection{Perspectives related to recent mesh-invariant models}
In the recent literature, models based on differential operators acting on the mesh, such as DiffusionNet \cite{sharp2022diffusionnet} and DeltaConv \cite{Wiersma2022DeltaConv} shows promising results with reassuring theoretical facts. However, these models have still not been applied to generative tasks and may require additional work but will be investigated in the future. In particular, their reliance on the intrinsic properties of a surface, such as the Laplacian, makes them sensitive to noise and topology changes, as opposed to our PointNet-based auto-encoder. Our results for unsupervised generative tasks are thus competitive with the state-of-the-art as demonstrated by the experiences.  In particular, while there is still a gap between fully supervised and unsupervised methods, our work opens up the possibility of extending the amount of available data for supervised generative tasks. Towards this objective, the incorporation of mesh invariant deep learning models within our framework is a promising avenue of work to improve the expressiveness of such data. 
\section{Conclusion}

In this paper, we propose a novel deep learning based approach for face registration. We use a varifold representation of shapes and extend kernel metrics on varifold with a multi resolution kernel. Our asymmetric auto encoder allows to learn a map from meshes with variable discretization to a low dimensional latent space. We demonstrate that our method allows for an efficient registration of meshes, and the learned latent space allows for powerful and easy deformation on this dataset. 

In the future, we plan to extend this approach to new data, such as human bodies, or animals. But the most crucial work to do remains the development of a better encoder with a similar versatility than PointNet.
\section*{Acknowledgments}
This work is supported by the ANR project Human4D ANR-19-CE23-0020, and was further supported by Labex CEMPI (ANR-11-LABX-0007-01) and the Austrian Science Fund (grant no P 35813-N). The authors would also like to thank Alexandre Mouton (CNRS, UMR 8524 - LPP, Lille) and Deise Santana Maia (CNRS, UMR 9189 CRIStAL, Lille) for their advice and many fruitful conversations.
\appendix

\section{Proof of Theorem \ref{th:approx} on the independence to mesh structure of the varifold representation} \label{appendix}

Let $S$ be a smooth compact surface. Let $x\mapsto \vec n_S(x)$ be its outer normal vector field. There is a radius $r_S>0$ such that any $x$ point with distance less than $r$ from $S$ has a unique projection (i.e. closest point) on $S$, denoted $\xi(x)$. Note that $\xi(x)-x=x\pm d(x,S)\vec n_S(x)$. We have $r_S\leq1/\kappa_S$, with equality for most surfaces.

Let $f$ be a full triangle (corresponding to a face in a mesh) with its vertices belonging to a surface $S$, with center $c$, normal $\vec n_f$ and greatest edge length $\eta$. As long as $\eta$ is small enough, the projection $\xi:f\rightarrow S$ is one-to-one.
\begin{figure}[h!]
    \label{fig:approximation}
     \centering
     \includegraphics[width=0.4\linewidth]{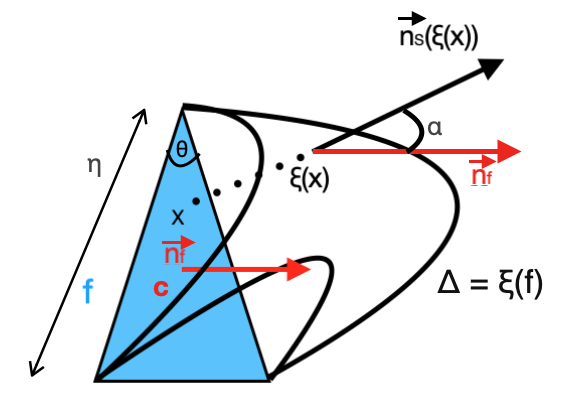}
     \caption{Face $f$ and its projection on a Surface $S$.}
 \end{figure}
We denote $\Delta=\xi(f).$ For $x$ in $f$, we have $d(\xi(x),x)\leq \eta\kappa_\Delta$, $\kappa_\Delta$ denotes the greatest eigenvalue of the second fundamental form of $S$ among all points of $\Delta$. See Figure A.16.

For any $y=\xi(x)$ in $\Delta,$ and $\eta$ small enough, we have (\cite[2.2.2]{MORVAN_closely_inscribed}). 
$$
d(y,c)\leq d(\xi(x),x)+d(x,c)\leq \eta\kappa_\Delta+\eta\leq (\kappa_\Delta+1)\eta.
$$
On the other hand, let $\alpha_f=\max_{y\in \Delta} (\vec{n}_f,\vec{n}_S(y))$. 
Then we have, for any $y=\xi(x)$ in $\Delta$ \cite{approx_normal_vector_field}[Corollary 1],
$$
d(\vec n_S(y),\vec n_f)\leq \sqrt{2}\sin(\alpha)\leq 6\sqrt{2}\frac{\kappa_\Delta}{\sin\theta_f}\eta,
$$
with  $\theta$ the smallest of the three angles of $f$. Moreover, for $\eta$ small enough  \cite{approx_normal_vector_field}[Corollary 2],
$$
\vert a(f)-a(\Delta)\vert \leq 3a(\Delta)\kappa_\Delta\eta,
$$
with $a(\cdot)$ the area of a surface in $\mathbb{R}^3$.

Now take some bounded Lipschitz function $u:\mathbb{R}^3\times \mathbb{S}^2\rightarrow\mathbb{R}$ with Lipschitz constant $Lip(u)=k_u$. Let $\mu_\Delta$ be the varifold associated with the smooth surface $\Delta$ and $\mu_f=a(f)\delta_{c}^{\vec{n}_f}$ as in our paper. 

The main argument follows these two remarks, which are just the results above:
\begin{enumerate}
    \item for any point $y=\xi(x)$ in $\Delta$, the value of $u$ at  $(y,\vec{n}_S(y))$ is close to its value at $(c,\vec{n}_f)$. Indeed: $$
    \vert u(y,\vec{n}_S(y))-u(c,\vec{n}_f)\vert\leq k_u (d(y,c)+d(\vec{n}_S(y),\vec{n}_f)).
    $$
    We have estimates for both of these distances that go linearly to 0 as $\eta$ goes to 0, so that for any $C=6\sqrt{2}$,
    $$
    \vert u(y,\vec{n}_S(y))-u(c,\vec{n}_f)\vert\leq C
\frac{(\kappa_\Delta+1) }{\sin\theta_f}k_u \eta 
    $$
    \item the difference between the area of $f$ and that of $\Delta$ go linearly to $0$ as $\eta$ goes to $0$, so for any $C\geq 3$
    $$
    \vert (a(\Delta)-a(f))u(c,\vec{n}_f)\vert\leq Ca(\Delta)\kappa_\Delta\Vert u\Vert_{\infty}\eta,
    $$
    with $\Vert u\Vert_{\infty}=\sup_{(x,n)\in \mathbb{R}^3\times\mathbb{S}^2}\vert u(x,n)\vert$.
\end{enumerate}

Letting ourselves be guided by these remarks, we have 
$$
\mu_f(u)=a(f)u(c,\vec n_f)=\int_\Delta u(c,\vec n_f)d\sigma(x)+(a(f)-a(\Delta))u(c,\vec n_f).
$$ 
We therefore get, for $C$ big enough (e.g., C=10), that $\Vert \mu_\Delta(u)-\mu_f(u)\Vert$ is bounded by:
$$\begin{aligned}
&\int_\Delta \vert u(y,\vec n_S(y)-u(c,\vec n_f)\vert d\sigma(y)+\vert a(f)-a(\Delta)\vert u(c,\vec n_f)\\
&\leq C
\frac{(\kappa_\Delta+1)\eta k_u a(\Delta)}{\sin\theta_f}
+
Ca(\Delta)
\kappa_\Delta\eta 
\vert u(c,\vec{n}_f)\vert\\
&
\leq Ca(\Delta)\frac{(\kappa_\Delta+1)}{\sin\theta_f}(k_u+\Vert u\Vert_\infty)\eta.
\end{aligned}$$

Now if $X$ is a mesh inscribed in $S$ whose vertices are in $S$, such that $\xi$ is bijective from the full triangles of $X$ onto $S$, we can sum this estimate over all faces and get
$$
\vert \mu_S(u)-\mu_X(u)\vert\leq Ca(S)\frac{(\kappa_S+1)}{\sin\theta_X}(k_u+\Vert u\Vert_\infty)\eta,
$$
with $\kappa_S$ the greatest eigenvalue of the second fundamental form of $S$ (i.e. its greatest principal curvature), $\eta$ its greatest edge length and $\theta_X$ the smallest angle among all faces of $X$.

Finally, for two meshes $X,\hat{X}$ inscribed in $S$, a triangular inequality immediately gives
$$
\vert\mu_X(u)-\mu_{\hat{X}}(u)\vert\leq Ca(S)\frac{(\kappa_S+1)}{\sin\theta_X}(k_u+\Vert u\Vert_\infty)\eta,
$$
with $C=20$. Similar estimates allow the computation of kernel norms, with the addition of explicit bounds on $(k_u+\Vert u\Vert_\infty)$. Indeed, kernel norms are computed by integrating the kernel along the varifolds, and bounds on the Lipschitz constant of the kernel and its $\Vert\cdot\Vert_\infty$ are easily computed.

Note that, to keep the proof readable, the estimates were purposely rough, just to give an idea on the order of the convergence. In practice, edges are shorter and areas are smaller near high curvature areas, allowing much better approximation than suggested by the formula.



\end{document}